\documentclass[sigconf, screen, nonacm, dvipsnames]{acmart}

\usepackage[noend]{algorithmic}
\usepackage{algorithm}
\usepackage{multirow}
\usepackage{amsmath}
\usepackage{balance}
\usepackage{bbm}
\usepackage{booktabs}
\usepackage[inline]{enumitem} 
\usepackage{subcaption}
\usepackage{bm} 
\usepackage{xcolor}

\usepackage{hyperref}
\usepackage{lipsum}

\usepackage{wrapfig}

\usepackage{svg}

\usepackage{arydshln}
\makeatletter
\def\adl@drawiv#1#2#3{
        \hskip.5\tabcolsep
        \xleaders#3{#2.5\@tempdimb #1{1}#2.5\@tempdimb}%
                #2\z@ plus1fil minus1fil\relax
        \hskip.5\tabcolsep}
\newcommand{\cdashlinelr}[1]{%
  \noalign{\vskip\aboverulesep
          \global\let\@dashdrawstore\adl@draw
          \global\let\ adl@draw\adl@drawiv}
  \cdashline{#1}
  \noalign{\global\let\adl@draw\@dashdrawstore
          \vskip\belowrulesep}}
\makeatother

\DeclareMathOperator*{\argmax}{arg\,max}

\begin{document}

\title{Agentic Personalisation of Cross-Channel Marketing Experiences}

\author{Sami Abboud}
\thanks{Authors are listed alphabetically.}
\affiliation{
  \institution{aampe}
  \city{Paris}
  \country{France}
}
\author{Eleanor Hanna}
\affiliation{
  \institution{aampe}
  \city{Durham}
  \country{USA}
}
\author{Olivier Jeunen}
\affiliation{
  \institution{aampe}
  \city{Antwerp}
  \country{Belgium}
}
\author{Vineesha Raheja}
\affiliation{
  \institution{aampe}
  \city{Mumbai}
  \country{India}
}
\author{Schaun Wheeler}
\affiliation{
  \institution{aampe}
  \city{Cary}
  \country{USA}
}

\begin{abstract}
Consumer applications provide ample opportunities to surface and communicate various forms of content to users. From promotional campaigns for new features or subscriptions, to evergreen nudges for engagement, or personalised recommendations; across e-mails, push notifications, and in-app surfaces.
The conventional approach to orchestration for communication relies heavily on labour-intensive manual marketer work, and inhibits effective personalisation of content, timing, frequency, and copy-writing.
We formulate this task under a sequential decision-making framework, where we aim to optimise a modular decision-making policy that maximises incremental engagement for any funnel event.
Our approach leverages a Difference-in-Differences design for Individual Treatment Effect estimation, and Thompson sampling to balance the explore-exploit trade-off. We present results from a multi-service application, where our methodology has resulted in significant increases to a variety of goal events across several product features, and is currently deployed across 150 million users.
\end{abstract}

\maketitle

\section{Introduction \& Motivation}
Consumer-facing applications proactively communicate with users for a variety of reasons.
This includes notifying users of new or ongoing campaigns, nudging users to engage with certain product offerings, and promoting user-specific recommendations or promotional incentives.
These communication strategies are typically owned by Customer Relationship Management (CRM) teams, where marketing professionals are responsible for the optimisation and orchestration of messaging~\cite{Kumar2018}.
Messaging decisions are multiplex, varying across dimensions like message channel (e-mail, push notification, in-app content card), timing (both absolute and relative, e.g. close to exogenous events like limited-time sales), and desired outcome (general engagement, conversion, feature adoption).

Message effectiveness is dependent on these decisions.
Importantly, the optimal messaging strategy will depend on the end-user under consideration.
For historical reasons, CRM strategies typically rely on rule-based systems with coarse segmentation~\cite{Tynan1987}, severely limiting their ability to address between-user variability.
There is a clear need for user-level personalisation, but manual orchestration practices inhibit doing this effectively.

\begin{figure}[!t]
    \centering
    \includesvg[width=\linewidth]{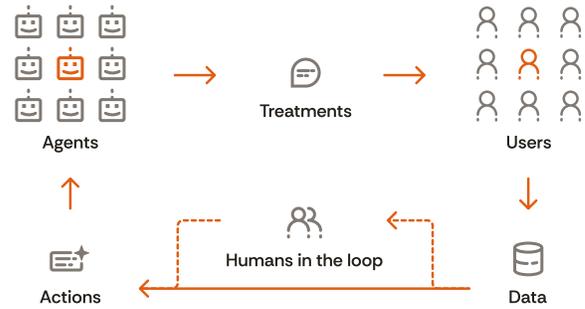}
    \caption{An illustration of the feedback loop in our setup. Logged user behaviour and their responses to interventions, together with a ``human-in-the-loop'' marketing professional, inform potential future messaging strategies, broken down to a modular action space. Agents personalise decision-making over these actions, balancing exploration with exploitation.}
    \label{fig:1}
\end{figure}

An additional hurdle to effective personalisation is that user preferences are naturally dynamic, whether due to psychological drift or as a function of previous app interactions.
Indeed---the effects of repetition and frequency can be multitude~\cite{Lee2014} and counter-intuitive~\cite{Kronrod2019,Batra1986}. 
This further compounds the computational load of manually orchestrated personalisation.

In this work, we describe a general methodology for scalable, dynamic, user-level personalisation of messaging strategies in an \emph{agentic} manner.
The \emph{agentic} framing underlines the focus on independent decision-making, and our methodology heavily draws from this literature~\cite{CONSEQUENCES2022}---leveraging methods from econometrics~\cite{McDowall2019}, causal inference~\cite{Athey2006}, and contextual bandits~\cite{Thompson1933,Russo2018}.
An \emph{agent} comprises an assembly of sequential decision-making modules, operating at the level of the individual user, across multiple dimensions of user experience and adapting over time in response to evolving user preferences~\cite{Trovo2020}.
Using a modified contextual bandit framework, an agent explores the user preference space by orchestrating user experiences, logging an individual user's behaviour in response to an experience, and balancing the exploitation of a preferred experience profile against continued exploration of the action space.
An agent cycles between implementing a user experience and analysing the result of its intervention (to inform its future user experience decisions).
Therefore, in this paradigm, CRM professionals have neither to implement user experience decisions themselves nor to perform an individual-level assessment of a user's preference, to achieve user-level personalisation; whilst retaining control over the action space that agents navigate.
Figure~\ref{fig:1} visualises this workflow.

We demonstrate the effectiveness of our approach to marketing communication personalisation through the results of a controlled experiment on messaging for users of a multi-service application, significantly increasing various goal events.

\section{Methodology \& Contributions}
Our goal is to personalise marketing communications between a consumer-facing application and its end-user customers.
This entails a range of decisions that need to be made: \emph{what} do we send to \emph{whom}, \emph{how} do we send it, and \emph{when}?
Some of these decisions can be thought of as modular in themselves: instead of treating every potential natural language message as part of an intractably large discrete action space, we can group attributes like tone-of-voice, emoji use, value propositions, recommendations, and promotional incentives into smaller and manageable action sets.
This enables us to tap into the existing decision-making and causal inference literature~\cite{CONSEQUENCES2022}, where recent advances extend classical methods to work with high-dimensional yet \emph{structured} treatments~\cite{Kaddour2021,Andreu2024}.

\paragraph{Measuring outcomes}
When actions are defined, we need to consider \emph{rewards}---the outcomes we wish to optimise~\cite{Jeunen2021Thesis}.
Importantly, we care about \emph{incrementality}, as we want a messaging strategy that drives outcomes, rather than merely correlating with them.
These ``outcomes'' are represented by event-stream data, which is high-dimensional and structured in and of itself.
Certain events will be considered \emph{goal} or \emph{end-of-funnel} events (like conversions or subscription renewals), whereas most are proxy events of varying importance (like home opens, page views, or clicks).

Let $Y^u_{[t_{i},t_{j}]}$ reflect the outcome for a user $u$, measured over a time period $[t_i,t_j]$.
From a dataset of timestamp and event tuples $\mathcal{D}_u = \{(t,e)_i\}_{i=1}^{n}$ for that given user, we can write:
\begin{equation}
    Y^u_{[t_{i},t_{j}]} = \sum_{(t,e) \in \mathcal{D}_u} w_{e}\cdot w_{t}.
\end{equation}
That is, the \emph{outcome} is a weighted sum of observed events for that user, with an event-specific weight $w_{e}$ and a temporal weight $w_{t}$.
Note that this formulation can easily encode metadata through $w_e$, such as order value, dwell time, or other attributes.

We model the event-specific weights as log-likelihood ratios on the conditional probabilities of (not) observing a goal event in a fixed window after said event: $w_e = \ln\left(\frac{\mathsf{P}(e|{\rm Goal})}{\mathsf{P}(e|\neg~{\rm Goal})}\right)$.
This indicates events' informativeness with respect to goal events in an intuitive manner---closely related to the concept of Granger causality~\cite{Shojaie2022}.

Temporal weights express that we are more interested in behaviour that happened closer to the intervention time $t_{\rm int}$.
We use an exponential decay with a tuneable half-life parameter.

To transform simple outcome measurements into Individual Treatment Effect (ITE) estimates, we leverage a sparse modification of Interrupted Time Series analysis~\cite{McDowall2019}.
We additionally account for periodic changes in \emph{organic} engagement, and hence, the method can more broadly be described as an instantiation of a Difference-in-Differences design~\cite{Athey2006}.
Suppose we observe a user who was subject to an intervention or treatment $T$ at time $t_{\rm int}$, and a user who was not subjected to it as control $C$, we define:
\begin{equation}
\Delta Y = \underbrace{(Y^{T}_{\rm post} - Y^{T}_{\rm pre})}_{\rm \Delta Y^{T}}-\underbrace{(Y^{C}_{\rm post} - Y^{C}_{\rm pre})}_{\rm \Delta Y^{C}}.
\end{equation}
This definition relies on one more hyperparameter $t_{\Delta}$, as the size of the time window to take into consideration:
\begin{equation}
    {\rm pre}=[t_{\rm int}-t_{\Delta}, t_{\rm int}) \quad\text{and}\quad{\rm post} =[t_{\rm int}, t_{\rm int}+t_{\Delta}).
\end{equation}

\paragraph{Controls}
In order to ensure that we measure \emph{causal} effects and not merely correlations between interventions and event frequency, we need to ensure that we correct for effects that are temporal or seasonal, and remove the variability that these effects bring.

That is, we define a group of users $C \subseteq \mathcal{U}$ that were not subject to the intervention to serve as control. There is a natural trade-off here. If we consider only the user that is closest to the individual whose ITE we are measuring, this would give us a low-bias but high-variance estimate. If, instead, we consider all possible users, we get high-bias but low-variance estimates. Ideally, we find a set of nearest neighbours to the treated user $T$ to consider.

\paragraph{Decision-Making}
For ITE estimates to inform decisions, we turn to the contextual bandit literature.
Specifically, we resort to Thompson sampling (TS)~\cite{Chapelle2011} as a stochastic arm selection heuristic with strong theoretical underpinnings~\cite{Jin2023}.
For a given context $x$ that describes the user's state, we sample an ITE estimate for every action ${\Delta{\tilde Y_{|a}}} \sim \hat{\mathsf{P}}(\Delta Y | A=a;X=x)$, selecting $\argmax_{a \in \mathcal{A}} \Delta{\tilde Y_{|a}}$.
Several methods can be used to estimate $\hat{\mathsf{P}}(\Delta Y |A;X)$.
When we view outcomes $Y$ as continuous and assume $\mathsf P(Y)$ to be Gaussian, $\Delta Y$ is also Gaussian (as a linear combination of Gaussians), which allows for closed-form updates.
Naturally, more advanced approaches like neural networks can also be plugged in to estimate ${\Delta{\tilde Y_{|a}}}$ directly.
We binarise $\Delta Y$ and use a Beta-Bernoulli model for interpretability.

We furthermore leverage an empirical Bayes prior to inform $\hat{\mathsf{P}}(\Delta Y |A;X)$ in the absence of data for a context-action pair, essentially imputing the distribution's parameters based on behaviour from similar users---inspired by user-based collaborative filtering~\cite{Ekstrand2011}.

\paragraph{Synthesising Messages}
Our framework relies on a modular design of the action space.
Aside from decisions about timing, frequency or communication channel, a message itself can consist of several actions (about tone-of-voice, offering and greetings, among others).
TS picks an action from all such action sets, and that combination of actions is then mapped back to the catalogue of potential messages a user is eligible for.
The best match gets sent to the user.
This modular decomposition can be framed as an instantiation of the Wolpertinger architecture that was designed for reinforcement learning with large discrete action spaces~\cite{DulacArnold2016}, and has recently seen adoption in recommender systems applications~\cite{SinghaRoy2023}.
Indeed, it allows us to avoid the combinatorial explosion of modelling every unique combination of individual actions as one.

\begin{table*}[!ht]
    \centering
    \caption{99\% confidence intervals for absolute increases in the percentage of the user-base engaging with upper- and bottom-funnel events for four important product features, over the control group. For GMV, we show a relative increase. Propensity score matching was applied to pair users who received at least one message with those in the control group who did not. }\label{tab:results}
    \begin{tabular}{lcclclclc}
    \toprule
       ~&~ &\textbf{Transactional PF 1} &~& \textbf{Transactional PF 2} &~& \textbf{Transactional PF 3} &~& \textbf{Account Creation PF} \\
    \midrule
    \textbf{Intent} &~& \colorbox{LimeGreen!50}{[+00.70\%, +00.83\%]} &~& \colorbox{LimeGreen!50}{[+01.76\%, +01.92\%]}&~& \colorbox{LimeGreen!50}{[+01.30\%, +01.44\%]} &~& \colorbox{LimeGreen!50}{[+02.37\%, +02.45\%]} \\
    \textbf{Conversion} &~& \colorbox{LimeGreen!50}{[+00.42\%, +00.48\%]}&~& \colorbox{LimeGreen!50}{[+00.30\%, +00.36\%]} &~& \colorbox{LimeGreen!50}{[+00.07\%, +00.13\%]}&~& \colorbox{LimeGreen!50}{[+00.21\%, +00.25\%]} \\
    \textbf{GMV} &~& \colorbox{LimeGreen!50}{[+14.12\%, +22.62\%]} &~& \colorbox{LimeGreen!50}{[+25.35\%, +43.39\%]} &~& \colorbox{Gray!50}{[--00.31\%, +22.55\%]} &~& N/A  \\
    \bottomrule
    \end{tabular}
\end{table*}

\paragraph{Human-in-the-loop}
Action sets relating to copywriting---such as tone-of-voice, value propositions, or calls-to-action---are maintained, updated and refined by marketing and product teams.
This ensures that businesses preserve control over what gets sent out, and professional human qualities can be leveraged to augment and amplify the system's potential performance.
Furthermore, it avoids problems of confabulation that run rampant with Large Language Models~\cite{Smith2023}, steering automated decision-making and enabling large-scale personalisation of communication, whilst respecting business guardrails by design.

\section{Empirical Results \& Discussion}
To empirically validate our proposed approach, we deploy a field experiment with a multi-service application spanning several markets.
The application supports various use-cases including ride-hailing and food delivery, providing a variety of product features and potential user experiences, which results in a rich test-bed for our proposed methodology.

The app engages users through a variety of messaging channels, including push notifications, in-app content cards, WhatsApp, email, and SMS.
These communications encompass promotional content, product marketing, user education, and transactional updates.
In this study, we evaluated an agentic messaging strategy within the push notification and in-app messaging channels, specifically targeting product marketing use cases.
The treatment group received agentically-orchestrated push notifications, which were layered on top of baseline rules-based messages delivered to both control and treatment groups.
Rules-based notifications were standardised in terms of timing and message copy, using heuristics optimised at the population level---the conventional approach.
In contrast, agentically-orchestrated messages were personalised with respect to timing, content, and delivery frequency.
For in-app messaging, the timing was contingent on triggering user behaviour within the app.
However, in the treatment condition, a subset of this content was personalised through agentic orchestration.

We validate our approach across four of the app's Product Features (PF).
The end-of-funnel event for three of these features was the completion of a transaction (e.g., completion of an order), with intent signals such as adding an item to cart.
For the fourth feature, the goal is to create an account for a specific service; intent signals included viewing the terms of service for the account or viewing the product description page.

\textit{Results.}
Table~\ref{tab:results} shows results from a three week randomised controlled experiment, across 6.4 million users.
We consider four key product features that drive engagement with the app, and are considered to be desirable from a business perspective.
For these features, we define three metrics
\begin{enumerate*}[label=(\roman*)]
    \item \textbf{Intent}, signifying the amount of users that shows interest in the feature by performing upper funnel events such as page views or clicks,
    \item \textbf{Conversion}, signifying the amount of users that make use of the feature and convert, and
    \item \textbf{Gross Merchandise Value (GMV)}, signifying the direct business impact of the conversions that were realised.
\end{enumerate*}

We observe significant improvements across both upper- and bottom-funnel events, across all considered use-cases.
Given the scale of the user-base, these results imply a significant practical business impact---highlighting the potential for our proposed methodology.
Today, the framework has been deployed and serves the entire user-base, supporting additional and evolving product marketing use-cases continually.
These results showcase the value of personalisation in marketing communication, a field that has historically largely focused on rule-based systems that would inhibit personalisation across all dimensions that matter.
Our proposed modularisation to accommodate a very large action space, combined with an extended contextual bandit framework that leverages adapted causal inference methods to interpret sparse and high-dimensional event-stream data, proves to be an effective approach.

More qualitatively, the agentic approach reduces the operational overhead of the orchestration stage significantly, and allows marketers to deploy an array of potential message variants far larger than what they would have been able to do manually.

\section{Conclusions \& Outlook}
The value of customer relationship management in consumer-facing applications is rarely underestimated.
Marketing communications serve as a key bridge between the business and its end-users.
Personalisation in this field is difficult because of the vast action space (i.e. natural language, timing, channels), and the vast amount of user attributes that influence messaging strategies' effectiveness over a multitude of business objectives.
For these reasons, the current status quo in practice consists of iterative A/B-testing of message variants and manual segmentation of the user-base, which amounts to a labour-intensive orchestration task that marketing professionals aim to surmount.
This leads to a clear bottleneck, and inhibits deeper personalisation of timing, copywriting, and beyond.

In this work, we propose a general agentic framework to tackle these problems, combining and adapting elements from several adjacent research areas: econometrics~\cite{McDowall2019}, causal inference~\cite{Athey2006}, and automated decision-making~\cite{CONSEQUENCES2022, Thompson1933}.
The resulting methodology allows for marketing professionals to deploy and scale a wide variety of communication strategies, personalised and optimised for incremental impact on various product-specific business metrics.

We report results from a field experiment on a large application, personalising marketing strategies for four different products.
Empirical observations underline the business value of the proposed approach, leading to significant improvements over all funnel metrics and double-digit relative improvements to GMV.
Our approach has been deployed to serve messages to the full user-base.

We note that the agentic approach does not serve to replace marketers, but rather augments their capabilities whilst operating autonomously and independently within the action space that is set up.
Agentic personalisation represents a practical, scalable step forward in how we design user-centric marketing systems.


\begin{acks}
    We would like to express our gratitude to Divyang Prateek Pandey and Saiyam Shah from the engineering team, whose expertise and dedication were instrumental in bringing our method from concept to reality. Their work in productionising the system, designing and implementing a robust, scalable infrastructure that delivers cross-channel personalisation at scale, was critical to serving real-world customers reliably and effectively. We also extend our sincere appreciation to Madhav Bhasin, Edward Keeble, Shashank Sahai, Chloe Meinshausen, Subhanshu Chaddha and Patricia Lazatin for their significant contributions in developing the end-to-end solution.
\end{acks}

\balance
\bibliographystyle{ACM-Reference-Format}
\bibliography{bibliography}


\begin{thebibliography}{21}


\ifx \showCODEN    \undefined \def \showCODEN     #1{\unskip}     \fi
\ifx \showDOI      \undefined \def \showDOI       #1{#1}\fi
\ifx \showISBNx    \undefined \def \showISBNx     #1{\unskip}     \fi
\ifx \showISBNxiii \undefined \def \showISBNxiii  #1{\unskip}     \fi
\ifx \showISSN     \undefined \def \showISSN      #1{\unskip}     \fi
\ifx \showLCCN     \undefined \def \showLCCN      #1{\unskip}     \fi
\ifx \shownote     \undefined \def \shownote      #1{#1}          \fi
\ifx \showarticletitle \undefined \def \showarticletitle #1{#1}   \fi
\ifx \showURL      \undefined \def \showURL       {\relax}        \fi
\providecommand\bibfield[2]{#2}
\providecommand\bibinfo[2]{#2}
\providecommand\natexlab[1]{#1}
\providecommand\showeprint[2][]{arXiv:#2}

\bibitem[Andreu et~al\mbox{.}(2024)]%
        {Andreu2024}
\bibfield{author}{\bibinfo{person}{Oriol~Corcoll Andreu}, \bibinfo{person}{Athanasios Vlontzos}, \bibinfo{person}{Michael O'Riordan}, {and} \bibinfo{person}{Ciaran~M. Gilligan-Lee}.} \bibinfo{year}{2024}\natexlab{}.
\newblock \bibinfo{title}{Contrastive representations of high-dimensional, structured treatments}.
\newblock
\newblock
\showeprint[arxiv]{2411.19245}~[stat.ML]


\bibitem[Athey and Imbens(2006)]%
        {Athey2006}
\bibfield{author}{\bibinfo{person}{Susan Athey} {and} \bibinfo{person}{Guido~W. Imbens}.} \bibinfo{year}{2006}\natexlab{}.
\newblock \showarticletitle{Identification and Inference in Nonlinear Difference-in-Differences Models}.
\newblock \bibinfo{journal}{\emph{Econometrica}} \bibinfo{volume}{74}, \bibinfo{number}{2} (\bibinfo{year}{2006}), \bibinfo{pages}{431--497}.
\newblock
\urldef\tempurl%
\url{https://doi.org/10.1111/j.1468-0262.2006.00668.x}
\showDOI{\tempurl}


\bibitem[Batra and Ray(1986)]%
        {Batra1986}
\bibfield{author}{\bibinfo{person}{Rajeev Batra} {and} \bibinfo{person}{Michael~L. Ray}.} \bibinfo{year}{1986}\natexlab{}.
\newblock \showarticletitle{Situational Effects of Advertising Repetition: The Moderating Influence of Motivation, Ability, and Opportunity to Respond}.
\newblock \bibinfo{journal}{\emph{Journal of Consumer Research}} \bibinfo{volume}{12}, \bibinfo{number}{4} (\bibinfo{year}{1986}), \bibinfo{pages}{432--445}.
\newblock
\showISSN{00935301, 15375277}
\urldef\tempurl%
\url{http://www.jstor.org/stable/254303}
\showURL{%
\tempurl}


\bibitem[Chapelle and Li(2011)]%
        {Chapelle2011}
\bibfield{author}{\bibinfo{person}{Olivier Chapelle} {and} \bibinfo{person}{Lihong Li}.} \bibinfo{year}{2011}\natexlab{}.
\newblock \showarticletitle{An Empirical Evaluation of Thompson Sampling}. In \bibinfo{booktitle}{\emph{Advances in Neural Information Processing Systems}}, \bibfield{editor}{\bibinfo{person}{J.~Shawe-Taylor}, \bibinfo{person}{R.~Zemel}, \bibinfo{person}{P.~Bartlett}, \bibinfo{person}{F.~Pereira}, {and} \bibinfo{person}{K.Q. Weinberger}} (Eds.), Vol.~\bibinfo{volume}{24}. \bibinfo{publisher}{Curran Associates, Inc.}
\newblock
\urldef\tempurl%
\url{https://proceedings.neurips.cc/paper/2011/file/e53a0a2978c28872a4505bdb51db06dc-Paper.pdf}
\showURL{%
\tempurl}


\bibitem[Dulac-Arnold et~al\mbox{.}(2016)]%
        {DulacArnold2016}
\bibfield{author}{\bibinfo{person}{Gabriel Dulac-Arnold}, \bibinfo{person}{Richard Evans}, \bibinfo{person}{Hado van Hasselt}, \bibinfo{person}{Peter Sunehag}, \bibinfo{person}{Timothy Lillicrap}, \bibinfo{person}{Jonathan Hunt}, \bibinfo{person}{Timothy Mann}, \bibinfo{person}{Theophane Weber}, \bibinfo{person}{Thomas Degris}, {and} \bibinfo{person}{Ben Coppin}.} \bibinfo{year}{2016}\natexlab{}.
\newblock \bibinfo{title}{Deep Reinforcement Learning in Large Discrete Action Spaces}.
\newblock
\newblock
\showeprint[arxiv]{1512.07679}~[cs.AI]


\bibitem[Ekstrand et~al\mbox{.}(2011)]%
        {Ekstrand2011}
\bibfield{author}{\bibinfo{person}{Michael~D. Ekstrand}, \bibinfo{person}{John~T. Riedl}, {and} \bibinfo{person}{Joseph~A. Konstan}.} \bibinfo{year}{2011}\natexlab{}.
\newblock \showarticletitle{Collaborative Filtering Recommender Systems}.
\newblock \bibinfo{journal}{\emph{Foundations and Trends® in Human–Computer Interaction}} \bibinfo{volume}{4}, \bibinfo{number}{2} (\bibinfo{year}{2011}), \bibinfo{pages}{81--173}.
\newblock
\showISSN{1551-3955}
\urldef\tempurl%
\url{https://doi.org/10.1561/1100000009}
\showDOI{\tempurl}


\bibitem[Jeunen(2021)]%
        {Jeunen2021Thesis}
\bibfield{author}{\bibinfo{person}{Olivier Jeunen}.} \bibinfo{year}{2021}\natexlab{}.
\newblock \emph{\bibinfo{title}{Offline Approaches to Recommendation with Online Success}}.
\newblock \bibinfo{thesistype}{Ph.\,D. Dissertation}. \bibinfo{school}{University of Antwerp}.
\newblock


\bibitem[Jeunen et~al\mbox{.}(2022)]%
        {CONSEQUENCES2022}
\bibfield{author}{\bibinfo{person}{Olivier Jeunen}, \bibinfo{person}{Thorsten Joachims}, \bibinfo{person}{Harrie Oosterhuis}, \bibinfo{person}{Yuta Saito}, {and} \bibinfo{person}{Flavian Vasile}.} \bibinfo{year}{2022}\natexlab{}.
\newblock \showarticletitle{CONSEQUENCES — Causality, Counterfactuals and Sequential Decision-Making for Recommender Systems}. In \bibinfo{booktitle}{\emph{Proceedings of the 16th ACM Conference on Recommender Systems}} (Seattle, WA, USA) \emph{(\bibinfo{series}{RecSys '22})}. \bibinfo{publisher}{ACM}, \bibinfo{address}{New York, NY, USA}, \bibinfo{pages}{654–657}.
\newblock
\showISBNx{9781450392785}
\urldef\tempurl%
\url{https://doi.org/10.1145/3523227.3547409}
\showDOI{\tempurl}


\bibitem[Jin et~al\mbox{.}(2023)]%
        {Jin2023}
\bibfield{author}{\bibinfo{person}{Tianyuan Jin}, \bibinfo{person}{Xianglin Yang}, \bibinfo{person}{Xiaokui Xiao}, {and} \bibinfo{person}{Pan Xu}.} \bibinfo{year}{2023}\natexlab{}.
\newblock \showarticletitle{Thompson Sampling with Less Exploration is Fast and Optimal}. In \bibinfo{booktitle}{\emph{Proc. of the 40th International Conference on Machine Learning}} \emph{(\bibinfo{series}{Proc. of Machine Learning Research}, Vol.~\bibinfo{volume}{202})}. \bibinfo{publisher}{PMLR}, \bibinfo{pages}{15239--15261}.
\newblock
\urldef\tempurl%
\url{https://proceedings.mlr.press/v202/jin23b.html}
\showURL{%
\tempurl}


\bibitem[Kaddour et~al\mbox{.}(2021)]%
        {Kaddour2021}
\bibfield{author}{\bibinfo{person}{Jean Kaddour}, \bibinfo{person}{Yuchen Zhu}, \bibinfo{person}{Qi Liu}, \bibinfo{person}{Matt~J Kusner}, {and} \bibinfo{person}{Ricardo Silva}.} \bibinfo{year}{2021}\natexlab{}.
\newblock \showarticletitle{Causal Effect Inference for Structured Treatments}. In \bibinfo{booktitle}{\emph{Advances in Neural Information Processing Systems}}, \bibfield{editor}{\bibinfo{person}{M.~Ranzato}, \bibinfo{person}{A.~Beygelzimer}, \bibinfo{person}{Y.~Dauphin}, \bibinfo{person}{P.S. Liang}, {and} \bibinfo{person}{J.~Wortman Vaughan}} (Eds.), Vol.~\bibinfo{volume}{34}. \bibinfo{publisher}{Curran Associates, Inc.}, \bibinfo{pages}{24841--24854}.
\newblock
\urldef\tempurl%
\url{https://proceedings.neurips.cc/paper_files/paper/2021/file/d02e9bdc27a894e882fa0c9055c99722-Paper.pdf}
\showURL{%
\tempurl}


\bibitem[Kronrod and Huber(2019)]%
        {Kronrod2019}
\bibfield{author}{\bibinfo{person}{Ann Kronrod} {and} \bibinfo{person}{Joel Huber}.} \bibinfo{year}{2019}\natexlab{}.
\newblock \showarticletitle{Ad wearout wearout: How time can reverse the negative effect of frequent advertising repetition on brand preference}.
\newblock \bibinfo{journal}{\emph{International Journal of Research in Marketing}} \bibinfo{volume}{36}, \bibinfo{number}{2} (\bibinfo{year}{2019}), \bibinfo{pages}{306--324}.
\newblock
\showISSN{0167-8116}
\urldef\tempurl%
\url{https://doi.org/10.1016/j.ijresmar.2018.11.008}
\showDOI{\tempurl}


\bibitem[Kumar and Reinartz(2018)]%
        {Kumar2018}
\bibfield{author}{\bibinfo{person}{Vineet Kumar} {and} \bibinfo{person}{Werner Reinartz}.} \bibinfo{year}{2018}\natexlab{}.
\newblock \bibinfo{booktitle}{\emph{Customer relationship management}}.
\newblock \bibinfo{publisher}{Springer}.
\newblock


\bibitem[Lee et~al\mbox{.}(2014)]%
        {Lee2014}
\bibfield{author}{\bibinfo{person}{Pei Lee}, \bibinfo{person}{Laks~V.S. Lakshmanan}, \bibinfo{person}{Mitul Tiwari}, {and} \bibinfo{person}{Sam Shah}.} \bibinfo{year}{2014}\natexlab{}.
\newblock \showarticletitle{Modeling impression discounting in large-scale recommender systems}. In \bibinfo{booktitle}{\emph{Proceedings of the 20th ACM SIGKDD International Conference on Knowledge Discovery and Data Mining}} (New York, New York, USA) \emph{(\bibinfo{series}{KDD '14})}. \bibinfo{publisher}{ACM}, \bibinfo{address}{New York, NY, USA}, \bibinfo{pages}{1837–1846}.
\newblock
\showISBNx{9781450329569}
\urldef\tempurl%
\url{https://doi.org/10.1145/2623330.2623356}
\showDOI{\tempurl}


\bibitem[McDowall et~al\mbox{.}(2019)]%
        {McDowall2019}
\bibfield{author}{\bibinfo{person}{David McDowall}, \bibinfo{person}{Richard McCleary}, {and} \bibinfo{person}{Bradley~J Bartos}.} \bibinfo{year}{2019}\natexlab{}.
\newblock \bibinfo{booktitle}{\emph{Interrupted time series analysis}}.
\newblock \bibinfo{publisher}{Oxford University Press}.
\newblock


\bibitem[Russo et~al\mbox{.}(2018)]%
        {Russo2018}
\bibfield{author}{\bibinfo{person}{Daniel~J. Russo}, \bibinfo{person}{Benjamin~Van Roy}, \bibinfo{person}{Abbas Kazerouni}, \bibinfo{person}{Ian Osband}, {and} \bibinfo{person}{Zheng Wen}.} \bibinfo{year}{2018}\natexlab{}.
\newblock \showarticletitle{A Tutorial on Thompson Sampling}.
\newblock \bibinfo{journal}{\emph{Foundations and Trends® in Machine Learning}} \bibinfo{volume}{11}, \bibinfo{number}{1} (\bibinfo{year}{2018}), \bibinfo{pages}{1--96}.
\newblock
\showISSN{1935-8237}
\urldef\tempurl%
\url{https://doi.org/10.1561/2200000070}
\showDOI{\tempurl}


\bibitem[Shojaie and Fox(2022)]%
        {Shojaie2022}
\bibfield{author}{\bibinfo{person}{Ali Shojaie} {and} \bibinfo{person}{Emily~B. Fox}.} \bibinfo{year}{2022}\natexlab{}.
\newblock \showarticletitle{Granger Causality: A Review and Recent Advances}.
\newblock \bibinfo{journal}{\emph{Annual Review of Statistics and Its Application}} \bibinfo{volume}{9}, \bibinfo{number}{Volume 9, 2022} (\bibinfo{year}{2022}), \bibinfo{pages}{289--319}.
\newblock
\showISSN{2326-831X}
\urldef\tempurl%
\url{https://doi.org/10.1146/annurev-statistics-040120-010930}
\showDOI{\tempurl}


\bibitem[Singha~Roy et~al\mbox{.}(2023)]%
        {SinghaRoy2023}
\bibfield{author}{\bibinfo{person}{Aayush Singha~Roy}, \bibinfo{person}{Edoardo D'Amico}, \bibinfo{person}{Elias Tragos}, \bibinfo{person}{Aonghus Lawlor}, {and} \bibinfo{person}{Neil Hurley}.} \bibinfo{year}{2023}\natexlab{}.
\newblock \showarticletitle{Scalable Deep Q-Learning for Session-Based Slate Recommendation}. In \bibinfo{booktitle}{\emph{Proceedings of the 17th ACM Conference on Recommender Systems}} \emph{(\bibinfo{series}{RecSys '23})}. \bibinfo{publisher}{ACM}, \bibinfo{pages}{877–882}.
\newblock
\showISBNx{9798400702419}
\urldef\tempurl%
\url{https://doi.org/10.1145/3604915.3608843}
\showDOI{\tempurl}


\bibitem[Smith et~al\mbox{.}(2023)]%
        {Smith2023}
\bibfield{author}{\bibinfo{person}{Andrew~L. Smith}, \bibinfo{person}{Felix Greaves}, {and} \bibinfo{person}{Trishan Panch}.} \bibinfo{year}{2023}\natexlab{}.
\newblock \showarticletitle{Hallucination or Confabulation? Neuroanatomy as metaphor in Large Language Models}.
\newblock \bibinfo{journal}{\emph{PLOS Digital Health}} \bibinfo{volume}{2}, \bibinfo{number}{11} (\bibinfo{date}{11} \bibinfo{year}{2023}), \bibinfo{pages}{1--3}.
\newblock
\urldef\tempurl%
\url{https://doi.org/10.1371/journal.pdig.0000388}
\showDOI{\tempurl}


\bibitem[Thompson(1933)]%
        {Thompson1933}
\bibfield{author}{\bibinfo{person}{William~R. Thompson}.} \bibinfo{year}{1933}\natexlab{}.
\newblock \showarticletitle{On the Likelihood that One Unknown Probability Exceeds Another in View of the Evidence of Two Samples}.
\newblock \bibinfo{journal}{\emph{Biometrika}} \bibinfo{volume}{25}, \bibinfo{number}{3/4} (\bibinfo{year}{1933}), \bibinfo{pages}{285--294}.
\newblock
\showISSN{00063444}
\urldef\tempurl%
\url{http://www.jstor.org/stable/2332286}
\showURL{%
\tempurl}


\bibitem[Trovo et~al\mbox{.}(2020)]%
        {Trovo2020}
\bibfield{author}{\bibinfo{person}{Francesco Trovo}, \bibinfo{person}{Stefano Paladino}, \bibinfo{person}{Marcello Restelli}, {and} \bibinfo{person}{Nicola Gatti}.} \bibinfo{year}{2020}\natexlab{}.
\newblock \showarticletitle{Sliding-window thompson sampling for non-stationary settings}.
\newblock \bibinfo{journal}{\emph{Journal of Artificial Intelligence Research}}  \bibinfo{volume}{68} (\bibinfo{year}{2020}), \bibinfo{pages}{311--364}.
\newblock


\bibitem[Tynan and Drayton(1987)]%
        {Tynan1987}
\bibfield{author}{\bibinfo{person}{A.~Caroline Tynan} {and} \bibinfo{person}{Jennifer Drayton}.} \bibinfo{year}{1987}\natexlab{}.
\newblock \showarticletitle{Market segmentation}.
\newblock \bibinfo{journal}{\emph{Journal of Marketing Management}} \bibinfo{volume}{2}, \bibinfo{number}{3} (\bibinfo{year}{1987}), \bibinfo{pages}{301--335}.
\newblock
\urldef\tempurl%
\url{https://doi.org/10.1080/0267257X.1987.9964020}
\showDOI{\tempurl}


\end{thebibliography}
\end{document}